\documentclass[letterpaper]{article} 
\usepackage[preprint]{aaai2027}  
\usepackage[hyphens]{url}  
\usepackage{graphicx} 
\usepackage{natbib}  
\usepackage{caption} 
\usepackage{algorithm}
\usepackage{algorithmic}
\usepackage[table]{xcolor}
\usepackage{amssymb}
\usepackage{subcaption}
\usepackage{dsfont}
\usepackage{bm}
\usepackage{newfloat}
\usepackage{listings}
\DeclareCaptionStyle{ruled}{labelfont=normalfont,labelsep=colon,strut=off} 
\floatstyle{ruled}
\newfloat{listing}{tb}{lst}{}
\floatname{listing}{Listing}

\usepackage{booktabs}
\usepackage{amsmath}
\begin{document}
	%
\title{Benchmarking MLLMs via Cognitive Expected Scene Graph for Safety-Critical Visual Negation Understanding}
\author{
	Zhiyun Jiang\textsuperscript{\rm 1},
	Hanyong Wang\textsuperscript{\rm 1},
	Binbin Liang\textsuperscript{\rm 1},
	Yu Xie\textsuperscript{\rm 2},
	Menglong Yang\textsuperscript{\rm 1},
	Wei Li\textsuperscript{\rm 1}
}

\affiliations{
	\textsuperscript{\rm 1}Sichuan University, Chengdu, China\\
	\textsuperscript{\rm 2}Beijing Institute of Technology, Beijing, China\\
}

\maketitle
\begin{abstract}
	\begin{quote}
		True machine intelligence requires transcending passive pixel registration to master top-down functional reasoning over absent information via visual negation understanding. However, unconstrained visual negation paradigms remain overly open-ended, and pervasive affirmation bias causes both existing Multi-Modal Large Language Models (MLLMs) and evaluation metrics to fail under negative semantics. To solve these intertwined challenges systematically, we first anchor the boundaries of negation reasoning within specific cognitive goals. Specifically, by focusing on safety as a highly pragmatic and critical cognitive dimension, we define the task of \textbf{S}cene \textbf{N}egation \textbf{U}nderstanding under \textbf{S}afety Cognition (\textbf{SNUS}). Under this framework, we construct a high-fidelity negative caption dataset mapping dense assertions of localized hazards. Concurrently, we propose the Cognitive Expected Scene Graph (CESG) Score, a structure-grounded, polarity-aware evaluation metric. Extensive experiments demonstrate that while current models struggle on the task, traditional metrics completely collapse under semantic reversals. Conversely, our framework delivers a solid benchmark for SNUS, providing a rigorous foundation to advance risk-aware situational comprehension and counterfactual cognition.
	\end{quote}
\end{abstract}

\section{Introduction}

\noindent 
Modern scene understanding centers on perceiving, identifying, and localizing entities explicitly present in visual scenes. Driven by large-scale pre-training and alignment, modern multimodal large language models (MLLMs) master pixel-visible environments through affirmative captioning \cite{affirm_cap}. Consequently, scene understanding is fundamentally treated as an exercise in capturing visible reality.

\begin{figure}[!htbp]
	\centering
	\begin{subfigure}[t]{0.5\textwidth}
		\centering 
		\includegraphics[width=\linewidth]{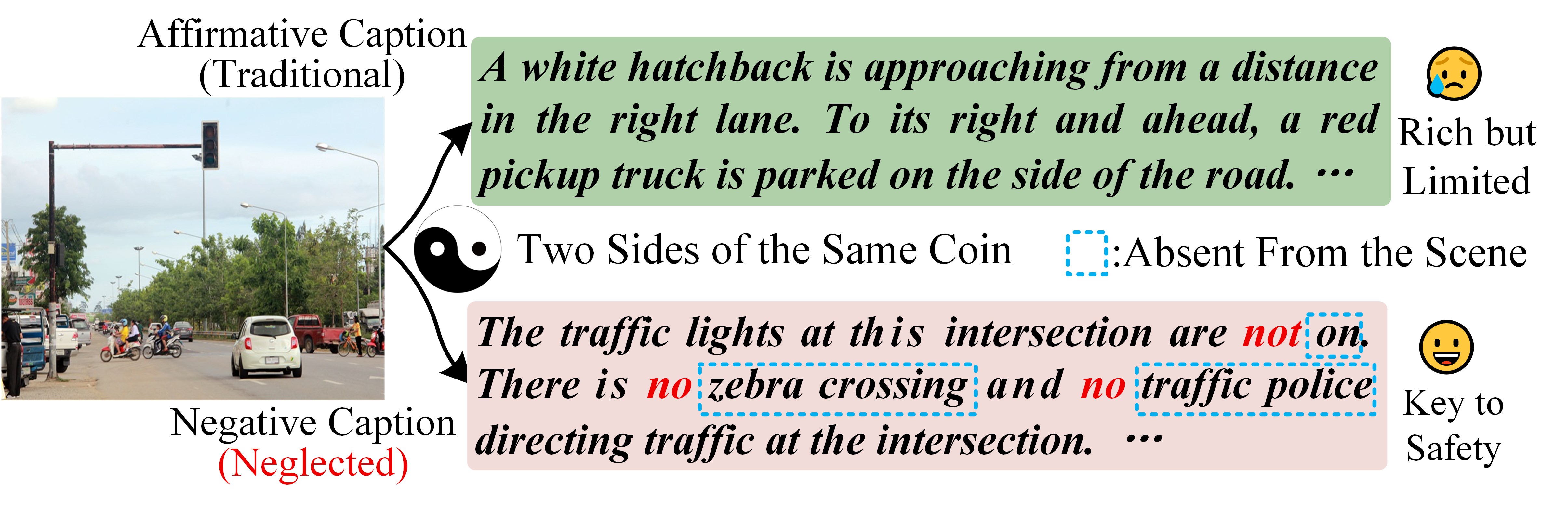}
		\caption{}
		\label{motivation1}
	\end{subfigure}
	\hfill
	\begin{subfigure}[t]{0.48\textwidth}
		\centering 
		\includegraphics[width=\linewidth]{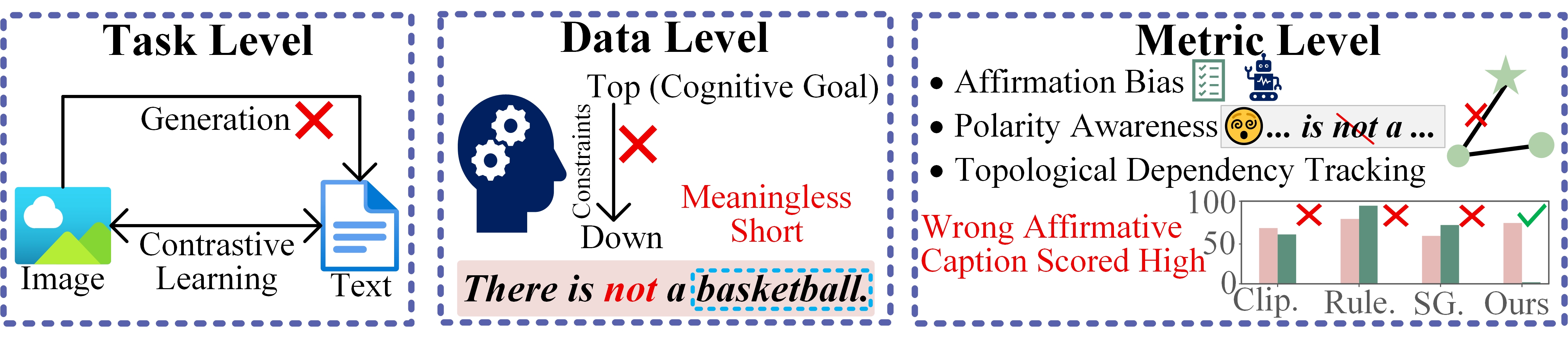}
		\caption{}
		\label{motivation2}
	\end{subfigure}
	\caption{Our Motivation. (a) Affirmative and negative captions under cognitive constraint constitute the two sides of the same coin. The latter is key to understanding risk, yet it is often overlooked. (b) Existing research is strictly constrained across task, data, and metric levels.}
	\label{motivation}
\end{figure}

However, a critical philosophical and functional shift is required to unlock true machine intelligence. The creation, maintenance, and evolution of visual scenes rely not only on explicit pixel-level information but also on certain absent information \cite{visual_dark}. Humans decode this absence through negation, a vital cognitive capability driving actionable functional understanding\cite{RS_Neg}. This capability becomes paramount in safety-critical domains like transportation or manufacturing, where the absence of key elements constitutes a semantic presence that directly defines risks, as shown in Figure \ref{motivation1}. To navigate these risks, machine perception must mimic this human capability, transcending passive recognition of affirmative content to master top-down functional reasoning over negative semantics.

Despite its necessity, existing research on visual negation \cite{valse,cc_neg,negbench,negrefcoco,NegVQA,medical_neg} is severely constrained by interrelated limitations across task, dataset, and metric dimensions, as shown in Figure \ref{motivation2}, ultimately limiting its practical utility in specialized domains. At the task level, existing negation paradigms primarily focus on contrastive learning between images and negative text, completely neglecting the generative captioning task. Consequently, at the data level, existing negative caption datasets lack high-level cognitive goal constraints, indiscriminately enumerating absent information that is mostly functionally meaningless (e.g., cognizing an absent \textit{basketball} is correct yet trivial). This contrastive focus also leaves a standardized negative captioning benchmark entirely vacant. Researchers thus inevitably resort to conventional metrics \cite{sc_captioner,SPICE,capture}, which inherit a strong affirmation bias \cite{affirmation_bias} and lack native polarity awareness, rendering them incapable of validating negative assertions. Furthermore, without topological dependency tracking to resolve structural ambiguities across objects, attributes, and relations, traditional evaluation metrics experience catastrophic collapse when encountering semantic reversals, coreference ambiguities, and cross-type semantic crossovers.

To address these interrelated challenges, we first formalize the task of Scene Negation Understanding under Safety Cognition (SNUS), bounding the negation reasoning scope to unlock practical utility. This formulation enables deep cognition of latent hazards in safety-critical domains, grounding high-level decision-making. At the data level, we introduce the SNUS dataset, the first high-quality negative long-caption corpus under safety constraints, systematically categorizing absent information across objects, attributes, and relations. By precisely mapping cognitive expectations to structured long-text captions comprising pure negative assertions, this corpus provides a rigorous testbed for fine-grained logical alignment. At the metric level, we propose the Cognitive Expected Scene Graph (CESG) Score. Unlike traditional scene graphs (SGs) mapping merely factual presence, the structurally grounded CESG inherently instills polarity awareness by explicitly encoding negative semantics into counterfactual object-attribute-relation topologies of visual scenes. Furthermore, to robustly track topological dependencies, CESG incorporates coreference resolution and precise semantic binding, preserving fine-grained structural dependencies even in complex settings. Finally, an integrated multi-stage pipeline assesses the three dimensions both individually and comprehensively, delivering a structure-grounded, polarity-aware evaluation engine. The synergy between the SNUS dataset and CESG Score establishes a novel methodological and evaluation foundation, ultimately guiding MLLMs toward deep scene cognition and top-down functional reasoning. Our contributions are summarized below:

\begin{itemize}
	\item We propose the novel task of visual scene negation understanding under cognitive constraints, guiding MLLMs to transcend passive pixel parsing toward deep cognition.
	\item With safety as the cognitive constraint, we construct the SNUS dataset, which uses purely negative captions to describe safety risk factors in scenes.
	\item We propose the CESG Score, which incorporates polarity awareness and topological dependency tracking, as the evaluation metric for the task.
	\item We systematically benchmark MLLMs, exposing current models' representational blind spots and conventional metrics' systemic limitations under negative semantics.
\end{itemize}

\section{Related work}
\subsection{Visual Negative Understanding}
Visual negative understanding aims to interpret absent scene elements. Early efforts rely on shallow perceptual mappings lacking high-level contextual reasoning \cite{helmet_detection_1,helmet_detection_2}. Recently, datasets like Valse\cite{valse}, CC-NEG\cite{cc_neg}, Negbench\cite{negbench}, NEGRefCOCOg\cite{negrefcoco}, NEG-TTOI\cite{negation1}, and CXR-Align\cite{medical_neg} have emerged to probe these capabilities. They typically formalize negative semantics via True Negatives (TN, absent in both images and text) or False Negatives (FN, present in images but described as absent). However, these paradigms predominantly focus on image-text contrastive matching rather than image-to-text generation, frequently utilizing FN to construct artificial counterfactual mismatches. Consequently, they lack dedicated metrics for negative captioning and suffer from data-level limitations, as shown in Table \ref{negation_dataset}. In contrast, to evaluate active deduction of real-world hazards over simple text-image contradictions, the SNUS dataset focuses exclusively on TN via long-form captions under explicit cognitive constraints. Paired with the CESG Score, this framework establishes a foundation for negative caption generation methodologies.

\begin{table}
	\centering
	\setlength{\tabcolsep}{1mm}
	\small
	\begin{tabular}{c|ccccccc}
		\hline
		\toprule
		
		\textbf{Dataset}&\textbf{Year}&\textbf{Length}&\textbf{Negation}&\textbf{Obj.}&\textbf{Att.}&\textbf{Rel.}&\textbf{CC} \\
		\midrule
		Valse &2022 & Short & TN, FN &  $\checkmark$ &$\times$ &$\times$ & $\times$ \\
		CC-Neg &2024 & Short & FN, FN  &$\checkmark$ &$\checkmark$ &$\checkmark$& $\times$ \\
		NegBench &2025 & Short & TN   &$\checkmark$ &$\times$ &$\times$ & $\times$\\
		NegRefCOCOg &2025 & Short & TN  &$\checkmark$ &$\checkmark$ &$\checkmark$  &$\times$\\
		NEG-TTOI &2025 & Short & TN   &$\checkmark$ &$\times$ &$\times$  & $\times$\\
		CXR-Align &2026 & Short & FN &$\checkmark$ &$\times$ &$\times$ &$\times$ \\
		
		\textbf{SNUS(Ours)} &2026 & Long & TN   &$\checkmark$ &$\checkmark$ &$\checkmark$&$\checkmark$ \\
		\bottomrule
	\end{tabular}
	\caption{Comparison of existing visual negation datasets.}
	\label{negation_dataset}
\end{table}

\subsection{Image Caption Evaluation}
Existing image captioning metrics broadly encompass lexical overlap, cross-modal embeddings, structural SGs, and emergent MLLM-based evaluators. Lexical metrics (e.g., BLEU \cite{bleu}) inevitably fail when a single missing negative word flips semantic polarity, while embedding-based approaches (e.g., CLIPScore \cite{Clipscore}) inherit a pervasive, systemic affirmation bias. Structural metrics (e.g., SPICE \cite{SPICE}, CAPTURE \cite{capture}, CompreCap \cite{comprecap}) utilize factual SGs to model semantic dependencies but ignore non-existence semantics and counterfactual reasoning. Meanwhile, MLLM-based paradigms \cite{mllm_cap_eval}, covering direct scoring and VQA prompting \cite{capability}, suffer from an acute sensitivity to model and prompt template variances, and the base models' inherent affirmation bias.

\section{SNUS Dataset}

\begin{figure}[!htbp]
	\centering
	\includegraphics[width=\linewidth]{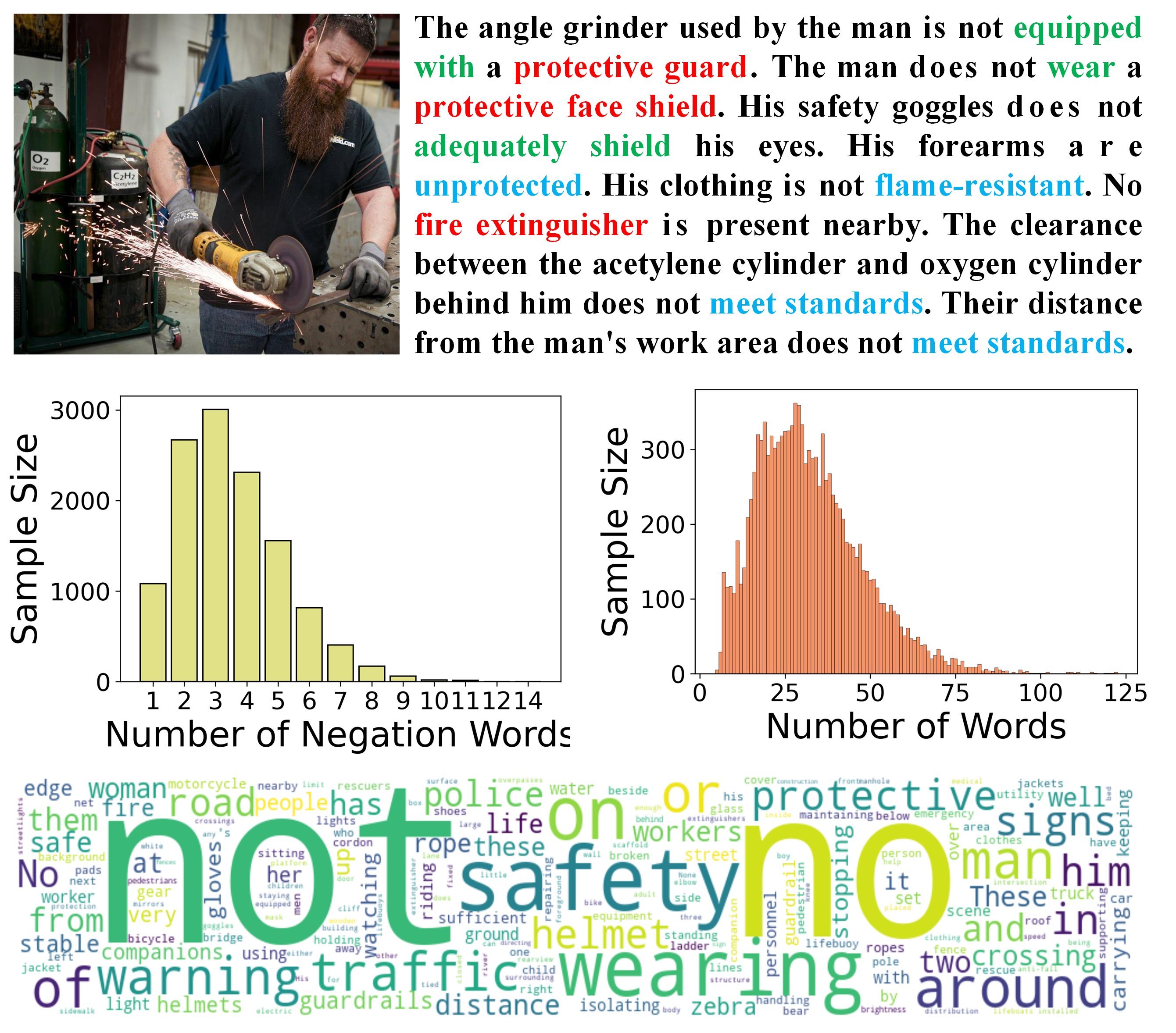}
	\caption{The first line shows a sample. We use red, green, blue to mark non-existent objects, relationships and attribute, respectively. The second line presents statistics on the number of negation words and the total number of words. The third line displays the word cloud of SNUS.}
	\label{data_statistics}
\end{figure}

\paragraph{Construction Pipeline}
We construct the dataset via a four-stage human-machine collaborative pipeline. First, we blend web-scraped and diffusion-generated safety-critical images to form a hybrid corpus. Next, GPT-4o\cite{gpt4o} automatically annotates missing information under safety constraints across objects (sub-divided into internal structural and external protective deficiencies), relationships, and attributes. MLLMs then synthesize these structured annotations into fluent, paragraph-level negative captions. Finally, a rigorous human-in-the-loop manual review eliminates model hallucinations, securing absolute ground-truth fidelity.

\paragraph{Data Statistics}

The SNUS dataset contains 12,118 image-caption pairs partitioned into train, validation, and test splits of 8,770, 1,548, and 1,800, respectively. Spanning diverse environments (e.g., roads, construction sites) and safety hazards (e.g., fire, fall), the corpus encapsulates 29,893 missing objects, 42,091 relationships, and 5,591 attributes (color-coded by category in Figure \ref{data_statistics}). Unlike prior benchmarks, SNUS features dense, multi-sentence paragraphs where every sentence embeds negation. Captions average 32.38 words (ranging from 5 to 122) and 3.51 negative words (ranging from 1 to 14) per entry, with detailed length and negation distributions illustrated in Figure \ref{data_statistics}.

\section{CESG Evaluation Framework}

\subsection{Challenge Analysis}
To overcome the limitations of linear token matching of rule-based metrics, a SG-based paradigm is highly desirable. It abstracts captions into structured topologies, preventing the alignment failures inherent in linear token streams. However, existing SG evaluation methods suffer from a series of challenges (\textbf{Q1}--\textbf{Q6}) that render them unsuitable for our task. Fundamentally, these challenges map directly to two core dimensions: polarity awareness (\textbf{Q1},\textbf{Q2}) and topological dependency tracking (\textbf{Q3}--\textbf{Q6}).

\paragraph{Polarity Awareness} 
Traditional SGs are strictly fact-driven and designed only to register physically present information. Consequently, they cannot represent or evaluate absent elements (\textbf{Q1}). Furthermore, they lack the cognitive capacity to verify the truth-value of correct negative claims omitted in the ground truth (\textbf{Q2}). Even though SC-Captioner\cite{sc_captioner} attempts to mitigate this, it remains confined to pixel-visible perception, failing to assess pixel-invisible information from a cognitive perspective.

\begin{figure*}[!htbp]
	\centering
	\begin{subfigure}[t]{0.73\textwidth}
		\centering 
		\includegraphics[width=\linewidth]{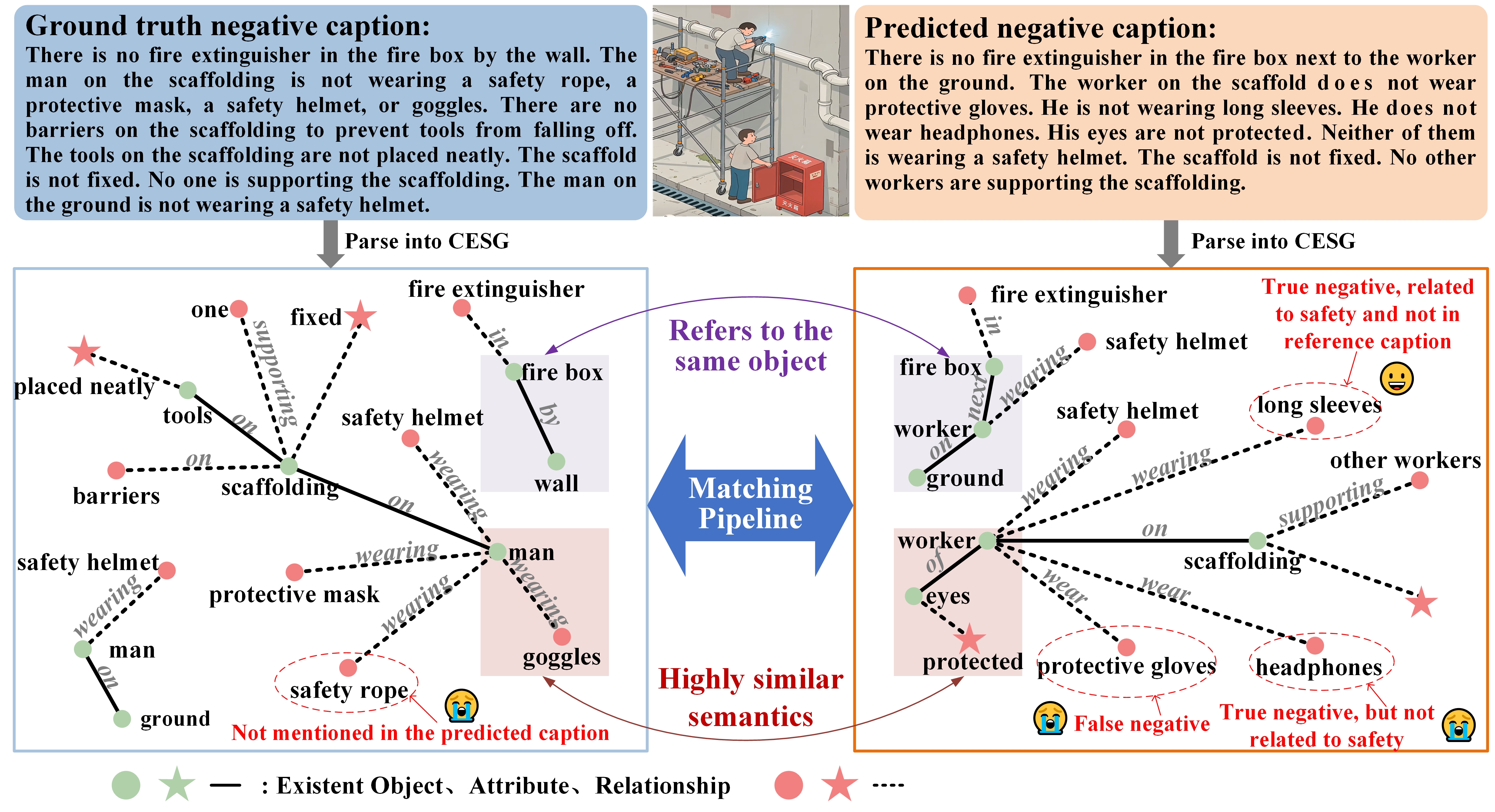}
		\caption{}
		\label{evaluation_overview}
	\end{subfigure}
	\hfill
	\begin{subfigure}[t]{0.26\textwidth}
		\centering 
		\includegraphics[width=\linewidth]{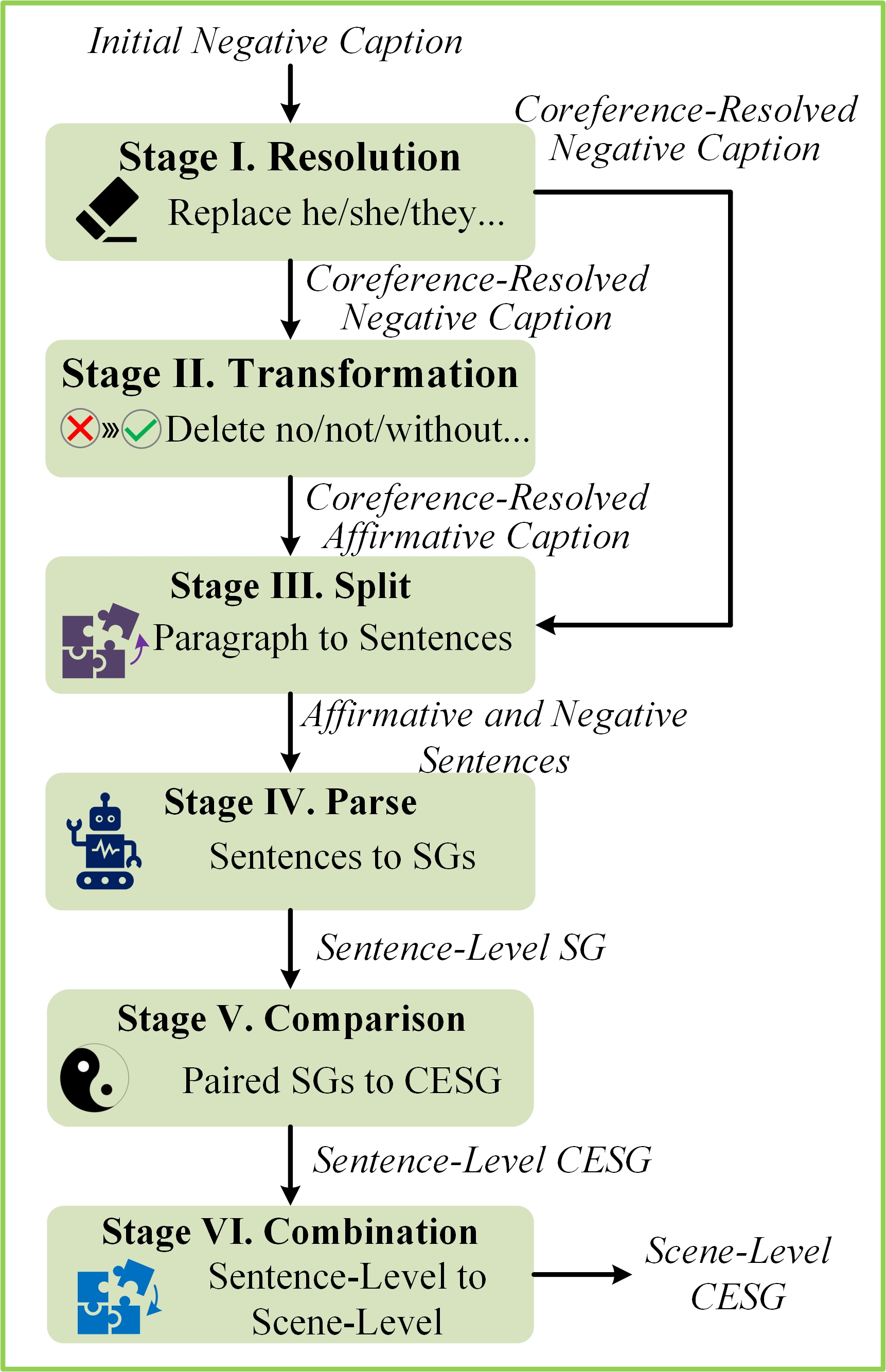}
		\caption{}
		\label{parse_process}
	\end{subfigure}
	\hfill
	\begin{subfigure}[t]{\textwidth}
		\centering 
		\includegraphics[width=\linewidth]{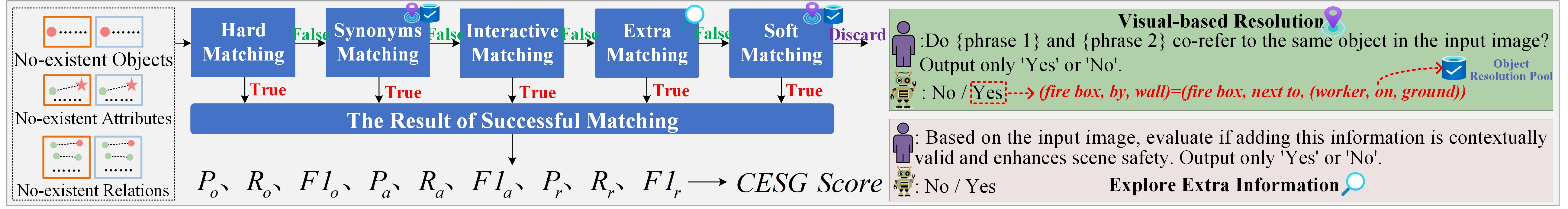}
		\caption{}
		\label{evaluation_process}
	\end{subfigure}
	\caption{CESG evaluation framework. (a) Overview. (b) CESG Construction Pipeline. (c) Evaluation Process.}
	\label{evaluation_illustration}
\end{figure*}

\paragraph{Topological Dependency Tracking} 
Existing methods fail to verify precise structural bindings, which is unacceptable in complex risky environments. They experience alignment collapse due to un-resolved coreferences where the same object is described differently across captions (\textbf{Q3}, e.g.,\textit{the fire box near the wall} and \textit{the fire box beside the worker on the ground} in Figure \ref{evaluation_overview} refer to the same object), and rigidly decouple semantically interchangeable relations and attributes (\textbf{Q4},  e.g., the semantic meaning of \textit{the man not wearing goggles} is highly similar to that of \textit{the worker's eyes are not protected}). Moreover, they cannot handle homogeneous description merging for plural objects (\textbf{Q5}, e.g., the prediction caption \textit{neither of these individuals is wearing a safety hat} semantically aligns with the reference captions describing each person individually in Figure \ref{evaluation_overview}), and fail to guarantee that negative attributes or relations are anchored to their specific, correct target objects (\textbf{Q6}, e.g., the non-existent safety helmet should be assigned to the worker standing on the scaffolding rather than the worker on the ground in Figure \ref{evaluation_overview}), which poses severe safety risks.

\subsection{Cognitive Expected Scene Graph}
Unlike vanilla SG, the CESG aligns with cognitive goals rather than the raw image (Figure \ref{evaluation_overview}). It may contain elements absent from the scene, while excluding task-irrelevant image details. Based on this goal-driven reconstruction, we evaluate non-existent information to address \textbf{Q1}.

\subsection{CESG Construction Pipeline}
We construct the CESG using DiscoSG-Refiner \cite{discosg}, which outperforms the Factual parser \cite{factual} (used in CAPTURE and SC-Captioner) and rivals GPT-4o \cite{gpt4o}. However, its direct application to negative captions suffers from three systemic flaws: (1) it overlooks negated attributes and relations, rendering edge polarities (solid vs. dashed) indistinguishable; (2) it fails to uniquely identify parallel homogeneous objects; and (3) it infers unstated relationships, introducing evaluation redundancy. To address these, despite its long-text capability, we adopt a six-stage pipeline of paragraph decomposition, sentence parsing, and graph synthesis as shown in Figure \ref{parse_process}.
\paragraph{Stage I: Resolution.} LINGMESS\cite{LingMess} is applied to resolve pronominal coreferences (e.g., \textit{he}, \textit{she}, \textit{they}) in negative captions, with plural pronoun resolution directly addressing the homogeneous description merging challenge (\textbf{Q5}).
\paragraph{Stage II: Transformation.} Stripping negation modifiers yields affirmative text. Despite potential syntactic anomalies, DiscoSG-Refiner's parsing remains unimpeded, rendering formal grammatical correction unnecessary.
\paragraph{Stage III: Split.} For multi-sentence inputs, the NLTK toolkit\cite{NLTK} tokenizes both the coreference-resolved negative captions and the transformed affirmative captions into individual sentence streams.
\paragraph{Stage IV: Parse.} DiscoSG-Refiner executes sentence-level parsing on the separated negative and affirmative sentences independently, generating a SG for each sentence.
\paragraph{Stage V: Comparison.} To counteract DiscoSG-Refiner's negation omissions, we cross-compare sentence-level negative and affirmative SGs. Because visible anchors manifest in both graphs while absent information is exclusive to the latter, this divergence isolates missing elements as black nodes and dashed edges to establish the CESG.
\paragraph{Stage VI: Combination.} Individual sentence-level CESGs are aggregated into a unified scene-level CESG under two rules. First, existing object nodes are merged and deduplicated if and only if their referential descriptions are identical. Second, absent information is strictly preserved without consolidation to avoid semantic loss.

\subsection{Evaluation Process}

Prior to calculating the CESG Score, we evaluate the precision, recall, and F1 score for each of the three types of missing information, following the workflow illustrated in Figure \ref{evaluation_process}. Here, precision measures the proportion of predicted information intersecting with the ground truth, while recall denotes the fraction of ground-truth information successfully retrieved. Because our evaluation focuses exclusively on pixel-level missing information, a filtering module first isolates subgraphs representing purely presence information, characterized by white nodes and solid edges, before the graphs enter the matching pipeline.

To address the structural dependency challenge (\textbf{Q6}), we define a unified dependency set $\mathcal{D}(e)$ for each predicted or ground-truth missing element $e$. Since non-existent attributes and relations are anchored to specific objects rather than isolated semantic units, flat-text evaluation causes severe misalignment. To prevent this, we explicitly extract the topological context of each missing element from the CESG. Let $\mathcal{I}^t = \mathcal{I}_p^t \cup \mathcal{I}_g^t$ represent the union of predicted and ground-truth sets of missing information for category $t \in \{o, a, r\}$\footnote{In subsequent evaluation notations, superscripts $o$, $a$, and $r$ denote objects, attributes, and relations, respectively, while subscripts $g$ and $p$ indicate the ground truth and predictions.}. Formally, the dependency set $\mathcal{D}(e)$ is defined hierarchically as a piecewise function:
\begin{equation}
	\mathcal{D}(e) = \begin{cases}
		\emptyset, & e \in \mathcal{I}^o \\
		\{(e_n, e_n^m)\}, & e \in \mathcal{I}^a \\
		\{(e_{sub}, e_{sub}^m), (e_{obj}, e_{obj}^m)\}, & e \in \mathcal{I}^r
	\end{cases}
\end{equation}
where $e_n$ represents the parent object for an attribute, and $e_{sub}$ and $e_{obj}$ denote the subject and object nodes for a relation, respectively. The superscript $m$ denotes their corresponding referential descriptions. For example, in the ground-truth CESG shown in Figure \ref{evaluation_overview}, for the non-existent relation triplet \textit{[fire extinguisher, in, fire box]}, $e_{obj}$ is \textit{fire box}, while $e_{obj}^m$ is \textit{[fire box, by, wall]}.

\paragraph{Hard Matching} At this stage, objects, attributes, and relations are matched independently. A match is only successful if both sides are completely identical, as shown below. 
\begin{equation}
	\begin{aligned}
		M_{h}(x,y) = \, & \mathds{1}\big[x = y \wedge \mathcal{D}(x) = \mathcal{D}(y)\big], \\
		& x \in \mathcal{I}_p^t, \; y \in \mathcal{I}_g^t, \; t\in\{o, a, r\}
	\end{aligned}
\end{equation}
where $M_h(\cdot)$ is the hard matching function, $\mathds{1}[\cdot]$ is the indicator function, and $x, y$ are elements to be matched from the prediction and ground truth, respectively.

\paragraph{Synonym Matching} Elements that fail hard matching enter this stage. Linguistic diversity necessitates verifying if unmatched elements are synonyms using WordNet $\mathcal{W}(\cdot)$. Furthermore, visually identical objects might possess disparate textual descriptions that evade synonym detection (\textbf{Q3}). To resolve this, we introduce a MLLM as a visual arbiter $Q_v(\cdot)$. A match is successful if the core elements are synonymous (or visually coreferent) and their dependency sets align either textually or visually:
\begin{equation}
	\begin{aligned}
		M_{s}(x,y) = \, & \mathds{1}\Big[ \big(x \in \mathcal{W}(y) \vee y \in \mathcal{W}(x) \vee Q_v(x, y)\big) \\
		& \wedge \mathcal{F}_{dep}\big(\mathcal{D}(x), \mathcal{D}(y)\big) \Big]
	\end{aligned}
\end{equation}
where $x \in \tilde{\mathcal{I}}_p^t, y \in \tilde{\mathcal{I}}_g^t, t \in \{o, a, r\}$, and $\tilde{\mathcal{I}}$ denotes the residual set from the previous stage\footnote{For simplicity, a uniform description is adopted across stages, despite variations in the remaining information.}. The function $\mathcal{F}_{dep}$ rigorously validates topological consistency, returning true if the elements in $\mathcal{D}(x)$ and $\mathcal{D}(y)$ are textually identical or deemed visually coreferent by $Q_v(\cdot)$.

\paragraph{Interactive Matching} To address semantic overlap across different information categories (\textbf{Q4}), we perform cross-category interactive matching. We utilize SimCSE\cite{SimCSE} to compute semantic similarity $f_s(\cdot, \cdot)$. To retain structural awareness, the element $e$ is concatenated with its dependency set $\mathcal{D}(e)$ to form a context-rich sequence $\mathbb{S}(e)$.  The matching function $M_{i}(x,y)$ operates across disparate categories:
\begin{equation}
	\begin{aligned}
		M_{i}(x,y) = \, & \mathds{1}\big[ f_s(\mathbb{S}(x), \mathbb{S}(y)) > \lambda \big], \\
		& x \in \tilde{\mathcal{I}}_p^t, \, y \in \tilde{\mathcal{I}}_g^{t^*}, \, t \neq t^* \in \{o, a, r\}
	\end{aligned}
\end{equation}
where $\lambda$ is a soft threshold. Embedding $\mathcal{D}(\cdot)$ into sequence $\mathbb{S}(\cdot)$ enforces strict topological priors to inherently penalize misaligned dependencies.

\paragraph{Extra Matching} This phase explicitly addresses information present in the prediction but absent from the ground truth (\textbf{Q2}). We leverage the MLLM $Q_e(\cdot)$ to evaluate whether integrating the residual predicted information enhances scene safety. To prevent hallucinations, the query seamlessly incorporates the topological anchor:
\begin{equation}
	\begin{aligned}
		M_{e}(x) = \mathds{1}\big[ Q_e(\mathbb{S}(x)) = \text{Yes} \big], x \in \tilde{\mathcal{I}}_p^t, \, t \in \{o,a,r\}
	\end{aligned}
\end{equation}

\paragraph{Soft Matching}
For any remaining elements within the same category that eluded prior discrete matches, we compute a continuous soft matching score matrix $\mathcal{S}^t$ to reward partial semantic alignments, analogous to the CAPTURE metric but structurally grounded:
\begin{equation}
	\begin{aligned}
		\mathcal{S}^t[i,j] =f_s\big(\mathbb{S}(\tilde{\mathcal{I}}_p^t[i]), \, \mathbb{S}( \tilde{\mathcal{I}}_g^t[j])\big), t\in\{o,a,r\}
	\end{aligned}
\end{equation}

\paragraph{Metric Calculation} 

Successful match sets from hard, synonym, interactive, and extra matching stages are denoted as $\mathcal{M}_h$, $\mathcal{M}_s$, $\mathcal{M}_i$, and $\mathcal{M}_e$, respectively, with superscripts distinguishing information types. Specifically, interactive matching ($\mathcal{M}_i$) employs dual superscripts to denote cross-category matches between prediction and ground truth. For example, $\mathcal{M}_i^{a,r}=\left\{(x,y)|x\in \tilde{\mathcal{I}_p^a},y\in \tilde{\mathcal{I}_g^r} \right\}$ pairs predicted attributes with ground-truth relations. Finally, precision, recall, and F1 scores are calculated as follows:
\begin{gather}
	P_{t} = \frac{|\mathcal{M}_h^t|{+}|\mathcal{M}_s^t|{+}|\mathcal{M}_e^t|{+}\sum\limits_{t^*\neq t}|\mathcal{M}_i^{t,t^*}|{+}\sum\limits_{i\in\mathcal{I}^t_p}\max_{j}\mathcal{S}^{t}[i,j] \ \ }{|\mathcal{I}_p^t|}, \\
	R_{t} = \frac{|\mathcal{M}_h^t|{+}|\mathcal{M}_s^t|{+}\sum\limits_{t^*\neq t}|\mathcal{M}_i^{t^*,t}|{+}\sum\limits_{j\in\mathcal{I}^t_g}\max_{i}\mathcal{S}^{t}[i,j] \ \ }{|\mathcal{I}_g^t|}, \\
	F1_{t} = \frac{2 \cdot P_t \cdot R_t}{P_t + R_t}, \quad t^*, t \in \{o, a, r\}
\end{gather}
To enable unified benchmarking, we introduce the overall CESG Score, inspired by CAPTURE's aggregation strategy. Instead of treating all categories equally, it is formulated as a weighted macro-F1 score:
\begin{equation}
	CESG \ Score= w_o \cdot F1_o + w_a \cdot F1_a + w_r \cdot F1_r
\end{equation}
where $w_o+w_a+w_r=1$.

\section{Benchmark Experiment}
\subsection{Experiment Settings}
\paragraph{MLLMs for Evaluation}
We benchmark 13 representative MLLMs across three paradigms: (1) \textbf{Commercial}: GPT-4o/5.4\cite{gpt4o,gpt54}, Claude-opus4.8/sonnet5\cite{claude_sonnet,claude_opus} and Gemini2.5/3.5-flash\cite{gemini_25,gemini_35}; (2) \textbf{Open-source}: LLaVA1.5-7B\cite{llava_15}, Llama3.2-11B\cite{llama32}, Qwen2.5V-3B/7B\cite{qwen25_VL}, Qwen3V-4B/8B\cite{qwen3}, and Intern3.5V-14B\cite{intern35}; and (3) \textbf{Supervised Fine-Tuning (SFT) Models}: the open-source backbones fine-tuned on our training split. Additionally, an Affirmative Caption (AC) baseline is constructed by stripping negation modifiers (e.g., \textit{no, not, without}) from ground truths to expose systemic affirmation bias.

\begin{table*}
	\setlength{\tabcolsep}{1mm}
	\small
	\centering
	\begin{tabular}{l | ccccccc | cccccccccc}
		\hline
		\toprule
		
		\textbf{Method}&\textbf{ROU}&\textbf{MET}&\textbf{CID}&\textbf{BL4}&\textbf{CAP}&\textbf{CS}&\textbf{SPI}&$\mathbf{P_o}$&$\mathbf{R_o}$&$\mathbf{F1_o}$&$\mathbf{P_r}$&$\mathbf{R_r}$&$\mathbf{F1_r}$&$\mathbf{P_a}$&$\mathbf{R_a}$&$\mathbf{F1_a}$ &\textbf{CESG} \\
		\midrule
		\multicolumn{18}{c}{\cellcolor{gray!18}\textbf{Commercial Models}} \\
		\midrule
		GPT-5.4& \textbf{21.44}& 29.67 &\textbf{2.05}&\textbf{5.87} & \textbf{34.40} &\textbf{29.59} &\textbf{11.10} & 75.30 & 34.70 & 47.51 & 72.31 & 25.26 &37.44 &68.12 &\underline{24.58} & \underline{36.12}&42.14 \\
		GPT-4o& 15.86& 27.45 &0.15 &3.52 & 33.47 &28.58 &8.76 &\textbf{92.29} & 34.79 & 50.53 & 82.22 & 25.36&38.77 &80.97 &23.58 &36.52 &44.09 \\
		Claude-sonnet5& 17.11& \textbf{31.57} &0.09 &4.46 & \underline{33.91} &\underline{29.35} &10.71 & 91.68 & \underline{45.14} & \textbf{60.49} & \underline{83.30} & \textbf{32.36}& \textbf{46.61} & \textbf{82.60} &19.35 &31.35 &\textbf{49.74} \\
		Claude-opus4.8& \underline{17.92}& \underline{30.63} &0.07 &\underline{4.95} & 33.35 &29.30 &\underline{10.63} & 78.40 & \textbf{45.40} & \underline{57.50} & 75.91 & \underline{27.05}&\underline{39.88} &68.12 &\textbf{26.68} &\textbf{38.34} &\underline{48.31} \\
		Gemini3.5-flash& 16.32 & 28.80 & 0.08 & 3.77 & 33.30 &28.25 &8.40 & \underline{92.03} & 37.03 & 52.81 & \textbf{84.81} & 22.11 &35.08 &80.76 &17.88 &29.28 &42.50 \\
		Gemini2.5-flash& 13.57 & 25.65 & \underline{0.52} & 2.53 & 31.87 &28.14 &6.86 & 77.79 & 28.79 & 42.02 & 74.92 & 26.44 &39.09 &73.29&3.21 &6.14 &32.32 \\
		
		\midrule
		\multicolumn{18}{c}{\cellcolor{gray!18}\textbf{Open-source Models}} \\
		\midrule
		LLaVA1.5-7B  & 17.19 & 15.16 & \textbf{3.69} & 2.87 & 30.60 &28.66 &\textbf{9.34} & 65.40 & 5.95 & 10.90 & 62.65 & 5.15 &9.52 &49.38 &3.09 &5.82 &9.29 \\
		Llama3.2-11B  & 9.80 & 17.34 & 0.30 & 0.77 & 30.35 &27.95 & 5.65& 72.57 & 23.67 & 35.69 & 64.16 & 22.32 &33.12 &56.16 &7.66 &13.48 &29.50 \\
		Intern3.5VL-14B  & 12.07 & \textbf{25.27} & 0.01 & 2.14 & \textbf{31.73} &\textbf{29.04} &6.15  & \textbf{87.59} & 33.06 & 48.00 & 74.95 & 25.75 &38.33 &77.08 &25.00 &37.75 &43.02\\
		Qwen2.5VL-3B  & 15.34 & 22.59 & 0.74 & 2.52 & 30.68 &28.42 &6.62  & 66.32 & 26.07 & 37.43 & 66.24 & 20.08 &30.82 &69.86 &1.27 &2.49 &27.04\\
		Qwen2.5VL-7B  & \textbf{18.77} & 22.69 & \underline{2.69} & \textbf{4.57} & 29.41 & 27.65 &\underline{8.60}  & 86.10 & 21.29 & 34.13 & \underline{78.86} & 13.59 &23.19 &\textbf{83.01} &13.86 &23.75 &28.80 \\
		Qwen3VL-4B  & 9.30 & 20.61 & 0.01 & 0.99 & 30.98 &27.80 &5.20   & \underline{86.67} & \underline{42.59} & \underline{57.11} & \textbf{81.31} & \underline{26.00} &\textbf{39.40} &73.94 &4.25 &8.04 & \underline{40.42} \\
		Qwen3VL-8B & \underline{15.34} & \underline{24.47} & 0.01 & \underline{2.33} & \underline{31.46} &\underline{27.79} &6.17   & 84.42 & \textbf{44.69} & \textbf{58.44} & 79.69 & \textbf{26.14} &\underline{39.37} &\underline{74.49} &\textbf{27.55} &\textbf{40.23} &\textbf{49.12} \\
		\midrule
		\multicolumn{18}{c}{\cellcolor{gray!18}\textbf{Supervised Fine-Tuning Models}} \\
		\midrule
		LLaVA1.5-7B  & \textbf{47.43}  & 50.80&\underline{106.87} & \textbf{34.87} & 41.30&22.35  &42.51& \textbf{95.40}& 51.13  & 66.58 & 84.10 & \textbf{25.94} & \textbf{39.65} & 81.77 & 25.00 & 38.29 & 52.78 \\
		Llama3.2-11B  & 47.13 & 50.65 & 105.93 & 34.33 & 42.02 &\textbf{27.57} &42.12  & 94.42 & \textbf{51.82} & \underline{66.91} & 84.18 & 25.92 &39.64 &81.25 &26.92 &40.44 &53.47  \\
		Intern3.5VL-14B  & 46.02 & 49.78 & 96.17 & 33.42 & 41.29 &27.46 &41.14& 94.57 & 50.76 & 66.06 & 84.58 & 25.19 &38.82 &76.44 &\textbf{29.97} &\textbf{43.06} &\underline{53.50} \\
		Qwen2.5VL-3B  & 46.54  & 49.66 & 103.56 & 33.83 & 40.94 & 22.34 & 41.29   & 94.08 & 50.62 & 65.82 & 84.06 & 25.57 & 39.22 & 79.21 & 27.57 & 40.91  & 52.94 \\
		Qwen2.5VL-7B  & 46.25 & 49.77 & 98.32 & 33.75 & \underline{41.48} &27.50 &41.19  & \underline{94.79} & 51.00 & 66.32 & 84.79 & 25.52 &39.23 &79.02 &27.37 &40.66 &53.13 \\
		Qwen3VL-4B  & 47.03 & 50.45 & 106.58 & 33.97 & 41.55 &27.51 &\underline{42.24}  & 94.04 & 51.08 & 66.20 & \underline{84.83} & \underline{25.77} &\underline{39.53} &\textbf{81.93} &\underline{28.92} &\underline{42.75} &\textbf{53.67} \\
		Qwen3VL-8B & \textbf{47.43} &  \textbf{51.17} & \textbf{111.20} & \underline{34.75} & \textbf{42.05} &\underline{27.62} &\textbf{42.78}  & 94.73 & \underline{51.90} & \textbf{67.06} & \textbf{85.43}& 22.51 &35.63 &\underline{81.55} &23.13 &36.03 &51.45  \\
		\midrule
		AC &91.64 & 90.04&63.90 & 71.72 & 59.41&21.66 &94.55 & 0 & 0 & 0 & 0  & 0 & 0 & 0 & 0 & 0  & 0\\
		\bottomrule
	\end{tabular}
	\caption{MLLM benchmarking results. Bold and underline indicate the best and second-best performance within each category. ROU: ROUGE-L; MET: METEOR; CID: CIDEr; BL4: BLEU-4; CAP: CAPTURE; CS: CLIPScore; SPI: SPICE.}
	\label{traditional_metrics}
\end{table*}

\paragraph{Implementation Details}
All experiments are conducted on 2 NVIDIA RTX A6000 GPUs. SFT is performed via the LLaMA-Factory framework\cite{llamafactory} using LoRA ($rank=8$)\cite{lora} with a learning rate of $1 \times 10^{-4}$ for 5 epochs. During inference, all MLLMs employ greedy decoding to ensure deterministic caption generation. For the CESG Score, the component weights are configured as $w_o = 0.5$, $w_a = 0.25$, and $w_r = 0.25$, with a comprehensive validation and discussion deferred to the ablation study.

\subsection{Metrics Comparison}
To evaluate the CESG Score, we benchmark it against traditional lexical (ROUGE-L\cite{rouge}, METEOR\cite{METEOR}, CIDEr\cite{cider}, BLEU-4\cite{bleu}), structural (CAPTURE\cite{capture}, SPICE\cite{SPICE}), and embedding-based (CLIPScore\cite{Clipscore}) metrics, demonstrating its superiority in visual negation understanding via two subsequent experiments.

\subsubsection{Robustness Against Affirmation Bias}
As shown in Table \ref{traditional_metrics}, the AC baseline serves as a rigorous stress-test to expose the vulnerability of evaluation paradigms to affirmation bias. Despite completely reversing the original negative semantics, AC counterintuitively achieves the highest performance across almost all traditional evaluation metrics, severely outperforming every commercial, open-source, and fine-tuned MLLM. Specifically, lexical metrics yield heavily inflated scores, with ROUGE-L, METEOR, and BLEU-4 peaking at 91.64, 90.04, and 71.72, respectively. Similarly, structural metrics collapse under this semantic shift, with CAPTURE and SPICE erroneously assigning their highest scores of 59.41 and 94.55 to the counterfactual AC text. This systemic failure occurs because traditional paradigms strictly reward surface-level token overlap or factual entity presence, remaining entirely blind to the fact that removing a single negative modifier completely invalidates the underlying safety context. Furthermore, although the embedding-based CLIPScore remains lower at 21.66, it still fails to effectively penalize this absolute semantic reversal because it functions as a visual bag-of-words matcher\cite{clip_bag_of_words} that is completely insensitive to negative logic.

Conversely, our framework exhibits absolute robustness against affirmation bias, assigning a score of 0 to the AC baseline across all fine grained dimensions and the final CESG Score. Since AC identifies no missing safety hazards, its counterfactual topology is entirely empty. This confirms that by explicitly modeling structures of non existence under cognitive constraints, the CESG Score resolves the representational blind spots plaguing traditional methods.

\begin{table}[t]
	\centering
	\setlength{\tabcolsep}{1mm}
	\small
	\begin{tabular}{l|cccccc}
		\toprule
		\textbf{Metric} & {\textbf{PCC }} & {\textbf{Cd $R^2$ }} & {\textbf{Kd $\tau$}} & {\textbf{Sa $\gamma$}}  & {\textbf{Sp $\rho$ }} & {\textbf{PA }}\\
		\midrule
		ROU & -0.4756 & 0.2262 & -0.3350 & -0.6941&-0.5228&0.1458 \\
		MET & -0.3529 & 0.1246 & -0.2583 & -0.6358&-0.4402&0.1816 \\
		CID & -0.5644 & 0.3185 & -0.3330 & -0.6765&-0.5120&0.1581 \\
		BL4 & -0.4017 & 0.1614 & -0.2940 & -0.6424&-0.4620&0.1716 \\
		CAP & -0.2032 & 0.0413 & -0.1740 & -0.4248&-0.2487&0.2694\\
		SPI & -0.6130 & \underline{0.3758} & -0.4778 & -0.7317 &-0.6520&0.1158\\
		CS & \underline{0.1092} & 0.0119 & \underline{0.0759} & \underline{0.1481} &\underline{0.1063}&\underline{0.5673}\\
		\textbf{CESG} & \bfseries 0.7578 & \bfseries 0.5743 & \bfseries 0.6336 & \bfseries 0.7866&\bfseries 0.7912&\bfseries 0.8641 \\
		\bottomrule
	\end{tabular}
	\caption{Assessment of the consistency between different evaluation metrics and human judgments}
	\label{metrics_evaluation}
\end{table}

\subsubsection{Consistency with Expert Judgement}
A reliable evaluation metric must align closely with human cognitive assessments. To evaluate alignment without prohibitive costs, we adopt the MLLM-as-a-Judge paradigm \cite{mllm_as_a_judge} via GPT-4o. Ground-truth captions are anchored at 8/10, reserving headroom (9–10) to credit models for discovering unannotated risk. We assess alignment using six statistical metrics. Pearson Correlation Coefficient (PCC) and Coefficient of Determination (Cd $R^2$) measure global linear correlation and absolute goodness-of-fit. Spearman’s Rank Correlation Coefficient (Sp $\rho$) and Kendall’s $\tau$ (Kd $\tau$) assess monotonicity and ranking consistency across the score spectrum, while Sample $\gamma$ (Sa $\gamma$) and Pairwise Accuracy (PA) evaluate fine-grained sample-level differentiation and pairwise discrimination accuracy within the same visual scene. The higher these statistical metrics are, the more the evaluation method aligns with human cognition.

As shown in Table \ref{metrics_evaluation}, traditional metrics fail catastrophically under negative semantics due to their systemic lack of negation awareness. Lexical metrics including ROUGE-L, METEOR, CIDEr, and BLEU-4 exhibit severe negative correlations, yielding a CIDEr PCC of -0.5644, Sp $\rho$ of -0.5120, Kd $\tau$ of -0.3330, Sa $\gamma$ of -0.6765, and abysmal PA values ranging from 14.58\% to 18.16\% because surface token matching ignores polarity flipping modifiers. Structural metrics such as CAPTURE and SPICE suffer an even steeper collapse, where SPICE records a PCC of -0.6130, Sp $\rho$ of -0.6520, Kd $\tau$ of -0.4778, Sa $\gamma$ of -0.7317, and a disastrous PA of 11.58\% despite a deceptive Cd $R^2$ of 0.3758, because factual SGs cannot represent latent counterfactual topologies of non-existence. Meanwhile, embedding-based CLIPScore delivers poor alignment across all dimensions, presenting a weak PCC of 0.1092, Sp $\rho$ of 0.1063, Kd $\tau$ of 0.0759, Sa $\gamma$ of 0.1481, and Cd $R^2$ of 0.0119, which culminates in a near random PA of 56.73\% due to its inherent affirmation bias as a visual bag of words matcher.

In contrast, our proposed CESG Score achieves state of the art alignment across all statistical dimensions, recording a commanding PCC of 0.7578, Cd $R^2$ of 0.5743, Kd $\tau$ of 0.6336, Sa $\gamma$ of 0.7866, Sp $\rho$ of 0.7912, and an impressive PA of 86.41\%. This compelling performance confirms that by integrating polarity aware topological matching with explicit coreference resolution, the CESG framework successfully overcomes affirmation bias and delivers a highly robust automated evaluation engine.

\subsubsection{Baseline Performance Analysis} 

Extensive benchmarking reveals that MLLMs severely struggle with visual negation. Under zero-shot settings, both open-source (Qwen3VL-8B: 49.12) and commercial models (Claude-sonnet5: 49.74) hit a distinct performance ceiling, falling short of fine-tuned baselines. Although SFT yields powerful improvements, allowing Qwen3VL-4B to peak at 53.67, a critical cognitive bottleneck persists. Crucially, overall performance is strictly bounded by recall rather than precision. For instance, while GPT-4o achieves an impressive 92.29\% object precision ($P_o$), its recall ($R_o$) stalls at 34.79\%, a bottleneck shared by SFT models whose recall hovers near 50\%. This pronounced precision-recall disparity originates from the pervasive affirmation bias in pre-training alignment, which predisposes MLLMs to passively map visible features instead of actively deducing counterfactual absences. Consequently, while models can accurately verify explicit hazards they notice, their scene cognition remains fundamentally incomplete due to a systemic failure in exhaustively positioning latent risks. Furthermore, the lower performance in relationships ($F1_r$) and attributes ($F1_a$) compared to objects ($F1_o$) underscores that reasoning over negated context requires a multi-hop logical chain because models must anchor the non-existent entity before resolving its structural dependencies.  

\subsection{Ablation Study}
\label{ablation}
\begin{table}[t]
	\centering
	\setlength{\tabcolsep}{1mm}
	\small
	\begin{tabular}{ccc| cccccc} 
		\toprule
		
		$w_{o}$ & $w_{a}$ & $w_{r}$ & \textbf{PCC} & \textbf{Cd ${R^2}$} & \textbf{Kd ${\tau}$} & \textbf{Sa ${\gamma}$ } & \textbf{Sp ${\rho}$} & \textbf{PA } \\
		\midrule
		1 & 0 & 0 & 0.7027 & 0.4938 & 0.6003 & 0.7861 & 0.7198 & 0.7034 \\
		0 & 1 & 0 & 0.5645 & 0.3187 & 0.5257 & \underline{0.7914} & 0.5993 & 0.5730 \\
		0 & 0 & 1 & 0.5642 & 0.3183 & 0.4843 & 0.6870 & 0.5885 & 0.6246 \\
		0 & 0.5 & 0.5 & 0.6583 & 0.4334 & 0.5544 & 0.7517 & 0.7013 & 0.7928 \\
		0.5 & 0 & 0.5 & 0.7292 & 0.5318 & \underline{0.6153} & 0.7655 & 0.7637 & 0.8117 \\
		0.5 & 0.5 & 0 & 0.7163 & 0.5131 & 0.6099 & \bfseries0.8052 & 0.7558 & 0.8076 \\
		1/3 & 1/3 & 1/3 & \underline{0.7443} & \underline{0.5540} & 0.6152 & 0.7791 & \underline{0.7757} & \underline{0.8588} \\
		0.5 & 0.25 & 0.25& \bfseries0.7578 & \bfseries0.5743 & \bfseries0.6336 & 0.7866 & \bfseries0.7912 & \bfseries0.8641 \\
		\bottomrule
	\end{tabular}
	\caption{Ablation study on the weights within CESG Score.}
	\label{weight_ablation_pure}
\end{table}

As shown in Table \ref{weight_ablation_pure}, we conduct a component ablation study to validate the weight configuration of the CESG Score. Relying solely on objects ($w_o=1$) establishes a strong baseline with a PCC of 0.7027 and a Sp $\rho$ of 0.7198. Eliminating object weights causes a catastrophic performance drop across all metrics, where the PCC and Cd $R^2$ collapse to approximately 0.56 and 0.31, respectively. This quantitative drop mathematically validates our cognitive hierarchy hypothesis, demonstrating that objects serve as the foundational topological anchors without which secondary attribute or relation annotations lose their contextual reference points. Conversely, while the object-only setting yields a limited PA of 0.7034, integrating attributes and relations boosts both the PA (0.8641) and Kd $\tau$ (0.6336). This improvement proves that incorporating $w_a$ and $w_r$ enforces strict topological penalties, which effectively prevents models from gaming the metric through simple entity word-matching when they misattribute negative edges to factual entities. Ultimately, our empirical configuration ($w_o=0.5, w_a=0.25, w_r=0.25$) achieves the peak performance among all evaluated configurations, balancing macro-level hazard discovery with micro-level structural dependencies.

\section{Conclusion}
We advance trustworthy scene cognition by formalizing the SNUS task. To overcome affirmation bias, we introduce a high-fidelity negative dataset mapping localized hazards, coupled with the CESG Score. By integrating polarity awareness with explicit topological dependency positioning, CESG resolves the evaluation collapse inherent in traditional metrics under semantic reversals. Extensive benchmarking confirms our framework establishes a rigorous evaluation standard for assessing top-down functional reasoning and trustworthy hazard positioning in safety-critical domains.

\bibliographystyle{aaai}
\bibliography{main}
\end{document}